\documentclass[10pt,twocolumn,letterpaper]{article}

 \usepackage{cvpr}              

\usepackage[dvipsnames]{xcolor}

\definecolor{cvprblue}{rgb}{0.21,0.49,0.74}
\usepackage[pagebackref,breaklinks,colorlinks,citecolor=cvprblue]{hyperref}

\def\paperID{2908} 
\def\confName{CVPR}
\def\confYear{2024}

\title{FurniScene: A Large-scale 3D Room Dataset with Intricate Furnishing Scenes}

\author{Genghao Zhang$^{1}$, Yuxi Wang$^{2,3}$, Chuanchen Luo$^{2}$, Shibiao Xu$^{1}$,\\ Zhaoxiang Zhang$^{2,3,4}$, Man Zhang$^{1}$, Junran Peng$^{2}$\\
{\normalsize\centering$^{1}$ Beijing University of Posts and Telecommunications}
{\normalsize\centering$^{2}$ Institute of Automation, Chinese Academy of Sciences}\\
{\normalsize\centering$^{3}$ Centre for Artificial Intelligence and Robotics, HKISI\_CAS, HongKong}\\
{\normalsize\centering$^{4}$ University of Chinese Academy of Sciences, UCAS}\\
{\tt\small \{ZGH, zhangman, shibiaoxu\}@bupt.edu.cn, \{luochuanchen2017,zhaoxiang.zhang\}@ia.ac.cn}\\
{\tt\small  yuxiwang93@gmail.com,} {\tt\small  jrpeng4ever@126.com}
}

\usepackage{indentfirst}
\begin{document}
\twocolumn[{%
\renewcommand\twocolumn[1][]{#1}%
\maketitle
\begin{center}
    \centering
    \captionsetup{type=figure}
    \vspace{-0.6cm}
    \includegraphics[width=0.98\linewidth]{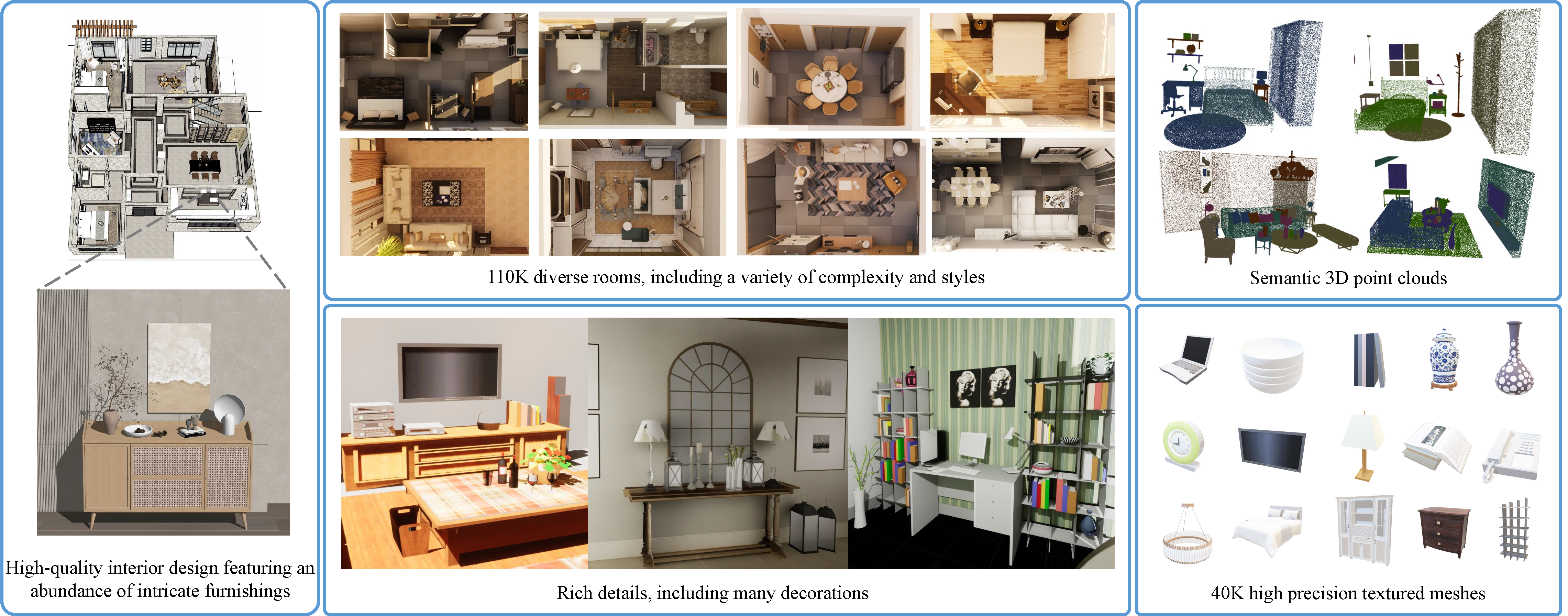}
    \vspace{-0.2cm}
  \caption{FurniScene is an extensive dataset comprising 111,698 meticulously crafted and diverse rooms, accompanied by 39,691 high-quality furniture CAD models. It encompasses 15 distinct room types and boasts a comprehensive collection of 89 object types, covering the majority of common items found in everyday life. In FurniScene, the furniture exhibits rich geometric details and high-resolution textures. Additionally, our dataset also includes a point cloud version and corresponding semantic annotations.}
  \label{fig:kaitou}
\end{center}%
}]

\begin{abstract}

Indoor scene generation has attracted significant attention recently as it is crucial for applications of gaming, virtual reality, and interior design. Current indoor scene generation methods can produce reasonable room layouts but often lack diversity and realism. This is primarily due to the limited coverage of existing datasets, including only large furniture without tiny furnishings in daily life. To address these challenges, we propose \textbf{FurniScene}, a large-scale 3D room dataset with intricate \textbf{Furni}shing \textbf{Scene}s from interior design professionals. Specifically, the FurniScene consists of 11,698 rooms and 39,691 unique furniture CAD models with 89 different types, covering things from large beds to small teacups on the coffee table. To better suit fine-grained indoor scene layout generation, we introduce a novel Two-Stage Diffusion Scene Model (TSDSM) and conduct an evaluation benchmark for various indoor scene generation based on FurniScene. Quantitative and qualitative evaluations demonstrate the capability of our method to generate highly realistic indoor scenes. Our dataset and code will be publicly available soon.

\end{abstract}    
\section{Introduction}
\label{sec:intro}

With the advent of content-generated techniques and large-scale datasets, recent years have witnessed rapid progress in indoor scenes generation~\cite{Procedural_generation_of_multistory_buildings_with_interior,Rule-based_layout_solving_and_its_application_to_procedural_interior_generation,ATISS}. The production of lifelike indoor scenes is imperative for applications such as game development, and virtual reality~\cite{yu2015clutterpalette,fisher2011characterizing}. Concurrently, the realms of interior design and decoration call for aesthetically pleasing and credible indoor scenes to meet the expectations of end-users~\cite{merrell2011interactive,yu2011make}.
To support research in indoor scene generation and understanding, data-driven approaches have attracted growing attention over the past few years~\cite{Fast-sync,talton2011metropolis,qi2018human,SUN3D,SceneNN,Matterport3D}.

Existing indoor scene datasets can be broadly categorized into three types based on the provided 3D semantic annotation: raw PCD, raw mesh~\cite{BuildingParser,SceneNN,Matterport3D,ScanNet}, and single mesh~\cite{Scan2CAD,OpenRooms,SceneNet,3D-FRONT}. Raw PCD and raw mesh datasets, such as BuildingParser~\cite{BuildingParser}, Matterport3D~\cite{Matterport3D}, and ScanNet~\cite{ScanNet} provide complexity and diversity scenarios as they are collected from real-world rooms. However, due to the lack of actual object positions and individual object meshes, they are not suitable for indoor scene layout generation. Alternatively, while 3D-FRONT~\cite{3D-FRONT} offers detailed 3D models of indoor scenes, placing a specific emphasis on furniture items, it does exhibit limitations in terms of diversity, particularly concerning scene variety and complexity. This is primarily reflected in the limited variety of furniture types, including only 43 categories of objects. This leads to monotonous room layouts with fewer objects, deviating from real-life scenarios.

To achieve real-life indoor scene generation, we present FurniScene, a novel large-scale 3D room dataset consisting of intricate furnishing scenes from interior design professionals. The proposed FurniScene showcases a striking level of diversity and intricacy, as shown in \cref{fig:kaitou}. Specifically, our dataset possesses the following four characteristics:  1) Meticulously designed interiors, curated directly by artists, ensure a high level of aesthetic quality; 2) The dataset boasts substantial data volume, encompassing 111,698 rooms and 39,691 individual CAD models, spanning 89 object types; 3) Diverse room furnishings perform as containing an average/maximum of 14.4/119 objects in each room; 4) Detailed scenes feature rich details, with each room fully furnished and containing numerous decor items. Additionally, our dataset encompasses point cloud data designed to facilitate research in 3D semantic segmentation. In \cref{tab:data_compare}, we display essential information for FurniScene and conduct a comparative analysis against various prominent indoor scene datasets.
\begin{table*}
  \centering
  \begin{tabular}{ccccccccc}
    \toprule
    Dataset & &  \#Rooms & \#Objects &  Texture & 3D Annotation & \#Object classes &  \#NOPM    \\
    \midrule
    SUN3D~\cite{SUN3D} & &  254 & N/A &  $\times$ & Raw PCD & N/R & N/A \\
    BuildingParser\cite{BuildingParser} & &  270 & N/A &  $\times$ & Raw PCD & 13 & N/A \\
    SceneNN~\cite{SceneNN} & &  101 & N/R &  \checkmark & Raw Mesh & N/R & N/A \\
    Matterport3D~\cite{Matterport3D} & &  2056 & N/A &  \checkmark & Raw Mesh & 40 & N/A \\
    ScanNet~\cite{ScanNet} & &  1506 & 296 &  \checkmark & Raw PCD/Mesh & N/A & N/A \\
    Scan2CAD~\cite{Scan2CAD} & &  1506 & 3049 &  $\times$ & Mesh & N/R & 9.3/40 \\
    OpenRooms~\cite{OpenRooms} & &  1068 & 2500 &  \checkmark & Mesh & 44 & N/A \\
    SceneNet~\cite{SceneNet} & &  57 & N/R &  $\times$ & Mesh & 40 & N/R \\
    3D-FRONT~\cite{3D-FRONT} & &  18,968 & 13,151 &  \checkmark & Mesh & 43  & 6.9/25\\
    \midrule
    FurniScene & &  \textbf{111,698} & \textbf{39,691} &  \checkmark & \textbf{Mesh/PCD} & 89 & \textbf{14.4/119} \\
    \bottomrule
  \end{tabular}
  \vspace{-0.2cm}
  \caption{Overview of representative 3D indoor scene datasets, where ``\#NOPM" represents the average/max number of objects per room, ``N/A"=``not available" and ``N/R"=``not reported"}
  \label{tab:data_compare}
\vspace{-0.4cm}
\end{table*}

Meanwhile, in order to address fine-grained indoor scene layout generation containing numerous decor items and diverse objects, we introduce a Two-Step Diffusion Scene Model (TSDSM) for indoor scene generation. In the first stage, we generate a furniture list specifying the category and size of each object present in the room. In the second stage, based on the furniture list from the first stage, we generate the complete layout information for the room. Because the first stage generates an accurate furniture list, the second stage does not produce an excessive or insufficient amount of furniture layout information, making the network easier to optimize. As a result, our method can generate more realistic and reliable rooms. Finally, based on FurniScene, we benchmark indoor scene generation tasks and evaluate the performance of several methods.

The contributions of this paper are summarized as follows. 1) We propose FurniScene, a novel synthetic indoor scene dataset with intricate furnishings, surpassing all existing datasets in both data volume and level of detail. Its primary advantage lies in the inclusion of numerous small furnishings, enhancing the realism of the scenes. 2) We propose TSDSM, a two-stage diffusion model for indoor scene generation, which demonstrated superior performance across multiple quantitative metrics. 3) We benchmark interior scene synthesis on FurniScene and provide qualitative and quantitative results, demonstrating its potential of promoting indoor scene generation.


\section{Related Work}
\label{sec:related_work}
\subsection{3D Indoor Scene Datasets}

In the past few decades, extensive indoor scene datasets have been proposed~\cite{SUN3D,BuildingParser,SceneNN,Matterport3D,ScanNet,Scan2CAD,OpenRooms,SceneNet,Structured3D,3D-FRONT}. These datasets can be broadly classified into three categories based on the types of 3D information they provide: raw PCD, raw mesh, and single mesh.

\textbf{Raw PCD and Raw Mesh Dataset}~~To construct a Raw PCD/Mesh type dataset, researchers capture RGB-D videos and utilize computer vision techniques to reconstruct scene point clouds or meshes. Subsequently, they manually add semantic information to the reconstructed data. SUN3D~\cite{SUN3D} includes 415 video sequences collected from 254 distinct rooms, with 8 sequences containing corresponding reconstructed raw point clouds. SceneNN~\cite{SceneNN} comprises over 100 indoor scenes, all reconstructed as triangle meshes and annotated at both vertex and pixel levels. However, these datasets lack individual furniture CAD models, posing challenges in extracting layout information.

\textbf{Single Mesh Dataset}~~ScanNet~\cite{ScanNet} includes 1,506 RGB-D video streams of rooms, point-wise semantic annotations, reconstructed meshes, and a small number (296) of CAD models. Scan2CAD~\cite{Scan2CAD} adds more CAD models to ScanNet, increasing the number of furniture models to 3,049, making indoor scenes richer. OpenRooms~\cite{OpenRooms} includes 1,680 rooms with 2,500 individual models and is a scanned dataset. SceneNet~\cite{SceneNet} is created by indoor design artists and includes rich and exquisite indoor scenes, but unfortunately, it only contains 57 rooms. The aforementioned datasets exhibit limitations in terms of room quantity. 3D-FRONT~\cite{3D-FRONT} is large-scale indoor scene 3D datasets released in recent years, offering 18,968 rooms, along with 13,151 unique furniture objects, including transform information and semantic details for each piece of furniture within each room. However, 3D-FRONT includes a limited range of furniture categories (43), mainly focusing on larger items. Smaller decor items like books, flowerpots, cups, tableware, and photos are missing. This leads to significant disparities between the dataset and real-life scenes, resulting in a relatively monotonous appearance and a lack of diversity.

\subsection{Indoor Scene Generation}

Existing scene generation methods typically generate layout information for each piece of furniture, followed by retrieving objects from the furniture database based on this information and accurately placing them in the scene. These methods can be roughly categorized three classes: forward neural generative models~\cite{SG-VAE,End-to-end_optimization_of_scene_layout,SceneHGN,Scene_synthesis_via_uncertainty-driven_attribute_synchronization,Indoor_scene_generation_from_a_collection_of_semantic-segmented_depth_images,Deep_generative_modeling_for_scene_synthesis_via_hybrid_representations}, autoregressive methods~\cite{SceneFormer,ATISS,Learning_3d_scene_priors_with_2d_supervision,CLIP-Layout,Fast-sync,wang2019planit,wang2018deep}, and methods based on diffusion models~\cite{Lego-net,DiffuScene}.
%
Grains~\cite{Grains}, SG-VAE~\cite{SG-VAE}, and SceneHGN~\cite{SceneHGN} use AE or VAE to model the distribution of various attributes of furniture in the dataset to generate indoor scenes. However, the authenticity of scenes generated by these methods is difficult to guarantee. GAN-based methods~\cite{Indoor_scene_generation_from_a_collection_of_semantic-segmented_depth_images} can generate scenes with a single style, lacking diversity.
%
Fast-sync~\cite{Fast-sync}, SceneFormer~\cite{SceneFormer}, ATISS~\cite{ATISS}, and CLIP-Layout~\cite{CLIP-Layout} use autoregressive methods to solve the generation task. These methods generate the attributes of the next object at one forward process until all the objects in the entire room are generated. However, using an autoregressive approach to the generation task results in averaging over the entire dataset.
%
Currently, diffusion models~\cite{ddpm,song2019generative,song2020improved} have made great progress in many tasks, such as image generation~\cite{ldm,nichol2021glide,kim2022diffusionclip}, human pose estimation~\cite{holmquist2023diffpose,qiu2023learning}, voice synthesis~\cite{liu2022diffsinger,huang2022fastdiff}, \textit{etc.} LegoNet~\cite{Lego-net} addresses the scene rearrangement task, using a diffusion model to predict the correct pose for each object, however, it only takes into account the 2D information of the furniture and ignores the height information. DiffuScene~\cite{DiffuScene} represents the entire scene as a scene graph and employs multi-layer transformers and MLPs to predict noise on it, generating relatively realistic indoor scenes. However, this approach utilizes a denoising network for all attributes, making the optimization of the network challenging.

\section{FurniScene Dataset}
\label{sec:dataset_acquisition_framework}

\begin{figure*}[h]
  \centering
  \includegraphics[width=0.98\linewidth]{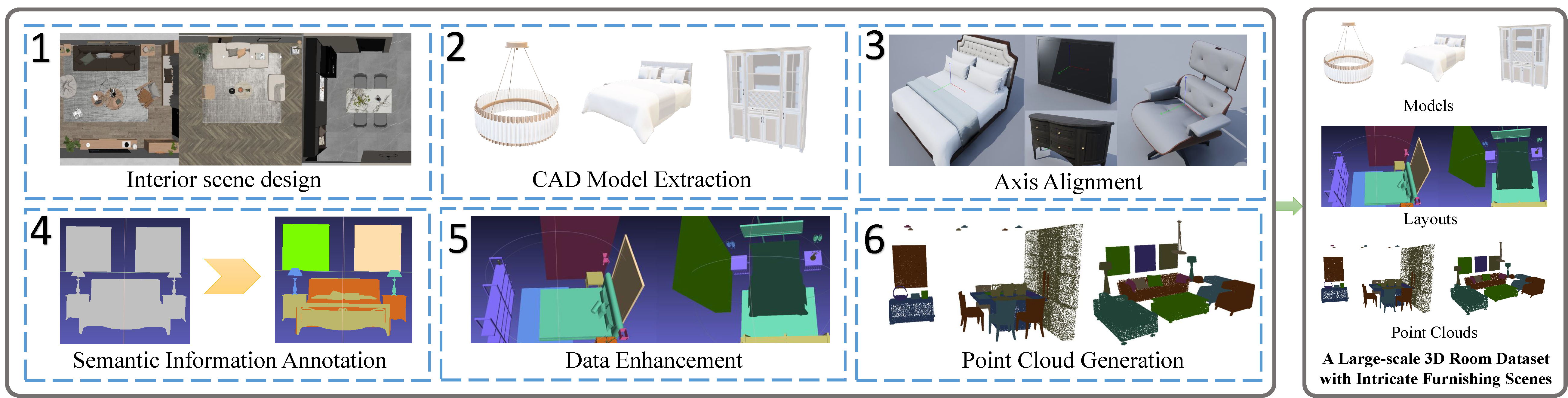}
  \vspace{-0.2cm}
  \caption{The pipeline of building FurniScene. Our data collection framework consists of the following steps: purchase complete SketchUp indoor scenes from interior designers; extract CAD model and align coordinate axis in 3DMax; render scenes in UE and add semantic labels manually; perform data augmentation to expand the dataset and increase scene diversity; point cloud generation.
 }
  \vspace{-0.4cm}
  \label{fig:pipeline}
\end{figure*}

\subsection{Dataset Acquisition Framework}

Creating a large-scale indoor scene dataset like FurniScene is a challenging task. Firstly, we purchase a large collection of meticulously designed SketchUp\footnote{\url{https://www.sketchup.com}} interior design assets from interior designers. Next, we import these assets into 3DMax\footnote{\url{https://www.autodesk.com/products/3ds-max}} for CAD model extraction and axis alignment. Subsequently, we transfer the processed assets into the Unreal Engine\footnote{\url{https://www.unrealengine.com}} (UE) using an intermediate format like UE's DATASMITH and perform semantic information annotation utilizing our in-house semantic labeling plugin. Finally, we performe data augmentation and point cloud generation. This is visually illustrated in \cref{fig:pipeline}.

\textbf{Interior Scene Design}~~The room layout information and furniture objects in our dataset are meticulously created by experienced professionals using SketchUp, a widely acclaimed 3D modeling software in architectural, interior decoration, and industrial product design. We select SketchUp for several reasons. Firstly, it is the software of choice for numerous interior design experts, so we can acquire a large quantity of professionally designed content. Furthermore, the data format utilized in SketchUp can seamlessly translate between industry-standard software like 3DMax and UE, greatly facilitating subsequent steps in data acquisition. To ensure data accuracy, we rigorously filter and exclude scenes with inaccuracies, such as excessive size or overlap, using methods similar to those employed in ATISS~\cite{ATISS} and DiffuScene~\cite{DiffuScene}.

\textbf{CAD Model Extraction}~~After obtaining the original room data, we identify two issues with the objects in the rooms: incomplete objects and instances where multiple objects are fused together. To address this, we utilize 3DMax, a professional modeling software that enables precise editing of each point and face of meshes. After importing the raw data into 3DMax, over twenty 3D modelers spent several months manually processing each object. Ultimately, we achieve precise mesh segmentation for each object.

\textbf{Axis Alignment}~~Next, we need to perform coordinate alignment for each object. Initially, we ensure that the coordinate axes of each object align with its center. Additionally, recognizing the pivotal role of furniture orientation in scene generation tasks~\cite{SceneFormer,ATISS,DiffuScene}, we process the coordinate orientations of each orientation-sensitive object, following 3D-FRONT~\cite{3D-FRONT}. The coordinate axes of each object are aligned according to the following rules: the axes follow a right-hand coordinate system, with the z-axis oriented upward, and the y-axis designating the front direction. After aligning, we obtain a complete and clear indoor scene.

\textbf{Semantic Information Annotation}~~After obtaining clean data through the preceding steps, we transfer it to UE with the aid of UE's DATASMITH. Then we need to utilize our in-house UE plugin for scene rendering and semantic annotation. Similar to OpenRooms~\cite{OpenRooms} and 3D-FRONT~\cite{3D-FRONT}, the semantic information of rooms in our dataset encompasses an array of vital attributes, including the bounding boxes for each room, furniture categories, orientation, size, and positional coordinates. Our UE data annotation plugin can automatically calculate the orientation, size, and position of each piece of furniture. Hence, data annotators only need to use our UE plugin to accomplish two tasks: in the top view of the room, use polygons to tightly wrap the entire room area and provide the category of the room; in the perspective view of the room, select each object and provide its semantic category. In this phase, we deliberately avoid imposing constraints on the object category labels. This approach enables us to extract a more diverse and comprehensive array of object categories, thereby enriching the dataset with a broader spectrum of semantic information.

\textbf{Data Enhancement}~~Similar to 3D-FUTURE~\cite{3D-FUTURE}, we use manually designed rooms as templates to create a greater variety of exquisite rooms to diversify the dataset. We apply three data augmentation methods: rotation, probabilistic deletion, and replacement. During the rotation operation, we introduce random angular adjustments of 90, 180, and 270 degrees for all furniture items within each room. To maintain the structural integrity of the rooms throughout deletion and replacement operations, we categorize items as deletable or replaceable. For deletable furniture, we apply deletion or replacement with a certain probability, while for replaceable furniture, we solely perform a replacement. We simultaneously apply these augmentation methodologies to each scene, amplifying the dataset by a factor of 100. Due to the large number of furniture items present in our template scenes and the extensive database of individual CAD models, there are not many similar rooms in our dataset after completing the data enhancement.

\textbf{Point Cloud Generation}~~To expand the applicability of FurniScene, we provide a point cloud data version aligned with semantic labeled point clouds in ScanNet~\cite{ScanNet}. Specifically, we evenly sample the mesh surface of each piece of furniture at 30,000 points, at the same time, assign each point a category label, as well as RGB color data strictly bound to the category, and finally output these data into a point cloud format.

\begin{figure*}[h]
  \centering
  \begin{subfigure}{0.32\linewidth}
    \includegraphics[width=1\linewidth]{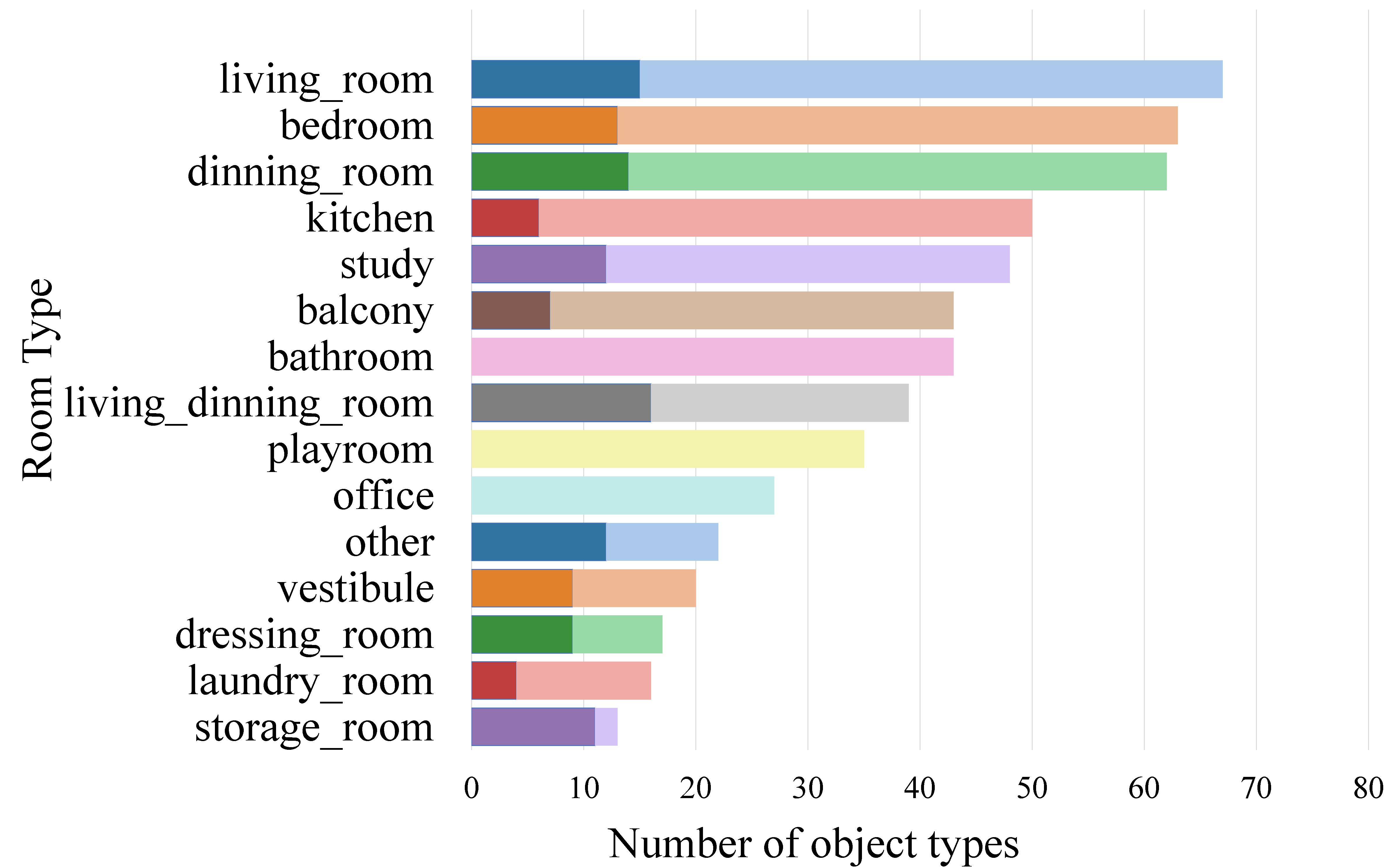}
    \caption{Total number of object types per room type.}
    \label{fig:short-a}
  \end{subfigure}
  \begin{subfigure}{0.32\linewidth}
    \includegraphics[width=0.98\linewidth]{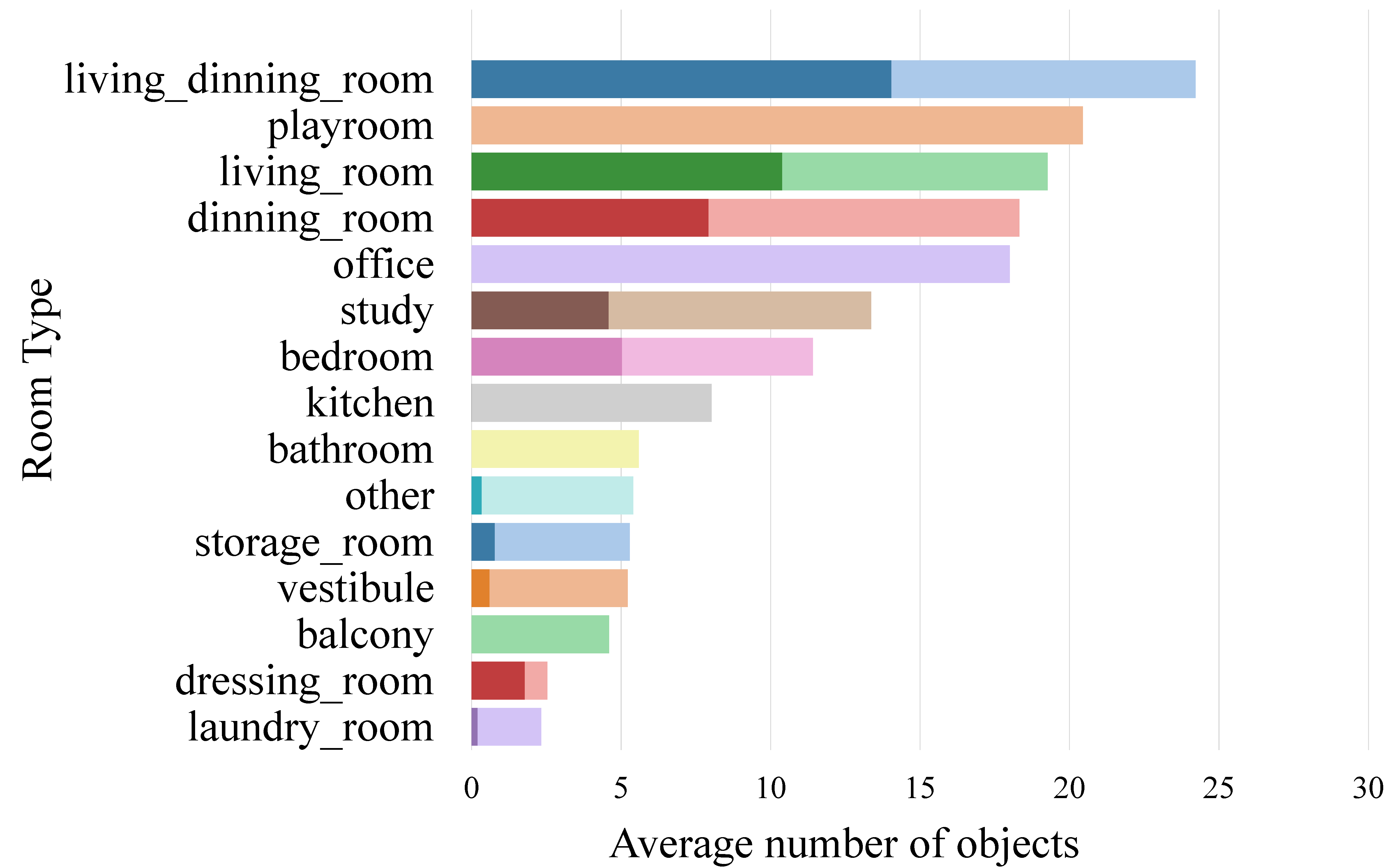}
    \caption{Average number of objects per room type.}
    \label{fig:short-b}
  \end{subfigure}
  \begin{subfigure}{0.32\linewidth}
    \includegraphics[width=0.98\linewidth]{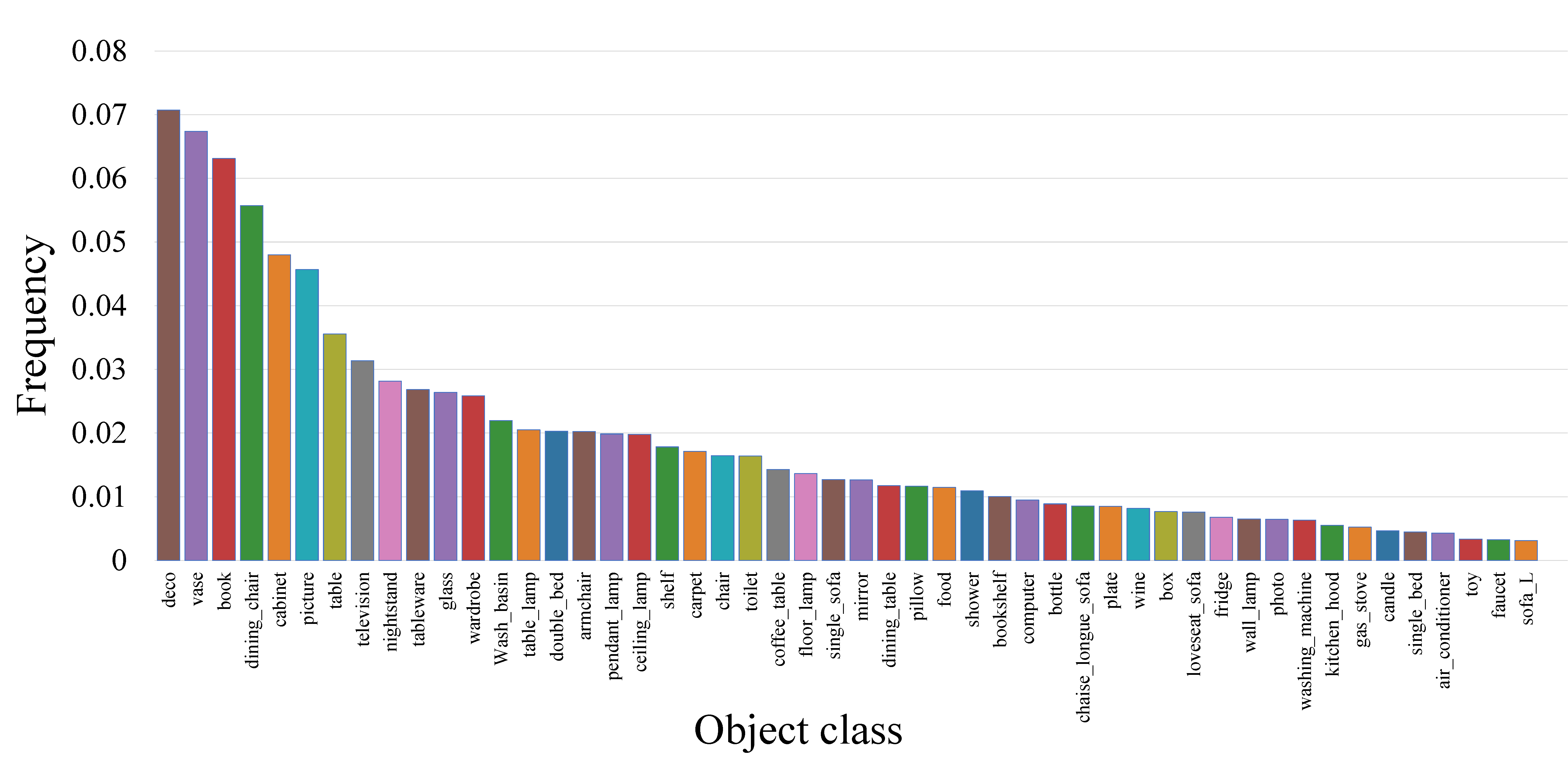}
    \caption{Objects Distribution.}
    \label{fig:short-c}
  \end{subfigure}
  \vspace{-0.2cm}
  \caption{Statistics on the dataset. The \textbf{darker} color represents the 3D-FRONT and the \textbf{brighter} color represents our FurniScene. a) shows the total count of object categories present in each room type, b) illustrates the average number of objects in each room type, and c) displays the distribution of the top 50 most frequently occurring objects in FurniScene.}
  \vspace{-0.4cm}
  \label{fig:statistics}
\end{figure*}

\subsection{Properties of FurniScene}

Compared to previous 3D datasets, FurniScene exhibits four prominent features that can better promote indoor scene 3D research: meticulous interior design, substantial data volume, diverse room furnishings, and rich details.

\textbf{Meticulous Interior Design}~~In FurniScene, interior designs are directly curated by artists. From the inception of the design process, our rooms are infused with rich details and meticulous aesthetic considerations. Furthermore, on furniture surfaces where objects can be placed, corresponding decorative items are thoughtfully included. The scene layouts exhibit meticulous organization, ensuring consistent styles, and many have already found practical applications.

\textbf{Substantial Data Volume}~~Our dataset contains 111,698 rooms designed by artists, many of which have been applied in practice. At the same time, there is little furniture reuse in FurniScene, so it includes a large amount of furniture CAD models (39,691). Most of these models come with corresponding high-resolution textures and high-precision geometric details. To the best of our knowledge, both the number of rooms in our dataset and the number of individual CAD meshes far exceed the existing dataset~\cite{Scan2CAD,OpenRooms,SceneNet,3D-FRONT}. 

\textbf{Diverse Room Furnishings}~~In the process of room design, existing datasets often feature relatively uniform styles, primarily including large furniture items such as bookshelves, coffee tables, bedside tables, and TV cabinets. This contrasts significantly with the complexity of real rooms. We intricately place various decorative items on these large pieces of furniture, such as vases on dining tables and cups on coffee tables. During the semantic information annotation process, we add category labels to these decorative objects in an open way. We uniformly process these decorative objects and ultimately obtain 40 categories of small decorative objects, which cover most household items, making our rooms more complete. Each type of room in FurniScene includes dozens of furniture types, far exceeding 3D-FRONT~\cite{3D-FRONT}, as shown in the \cref{fig:short-a}. Simultaneously, \cref{fig:short-c} highlights the extensive variety of decorative item categories included in FurniScene.

\textbf{Rich Details}~~The richness and authenticity of rooms are vividly depicted in FurniScene, marking one of the primary strengths of our dataset. To the best of our knowledge, our dataset exhibits the highest object density in each room among indoor scene datasets with object meshes. The rooms in our dataset contain an average of 14.4 objects, with a maximum of 119 objects, surpassing comparable datasets such as 3D-FRONT \cite{3D-FRONT} (average: 6.9 objects, maximum: 25 objects) and Scan2CAD \cite{Scan2CAD} (average: 9.3 objects, maximum: 40 objects). We present the comparison results for the average number of objects in each room type in \cref{fig:short-b}. 

\subsection{Data Annotation Costs}

Due to the elevated object density in each room, the data annotation process involves labor-intensive tasks, primarily encompassing CAD model extraction, axis alignment, and object category annotation. Annotating one room typically consumes 2-3 hours, with CAD model extraction demanding about 1 hour, axis alignment and category annotation requiring 1-2 hours. These tasks are completed by twenty students with 3D modeling knowledge. Furthermore, point cloud generation for a room is a CPU-intensive operation, taking approximately 30-60 seconds for each room. 
\section{Method}
\label{sec:method}
\subsection{Data Preparation}
\begin{figure*}[h]
  \centering
  \includegraphics[width=0.98\linewidth]{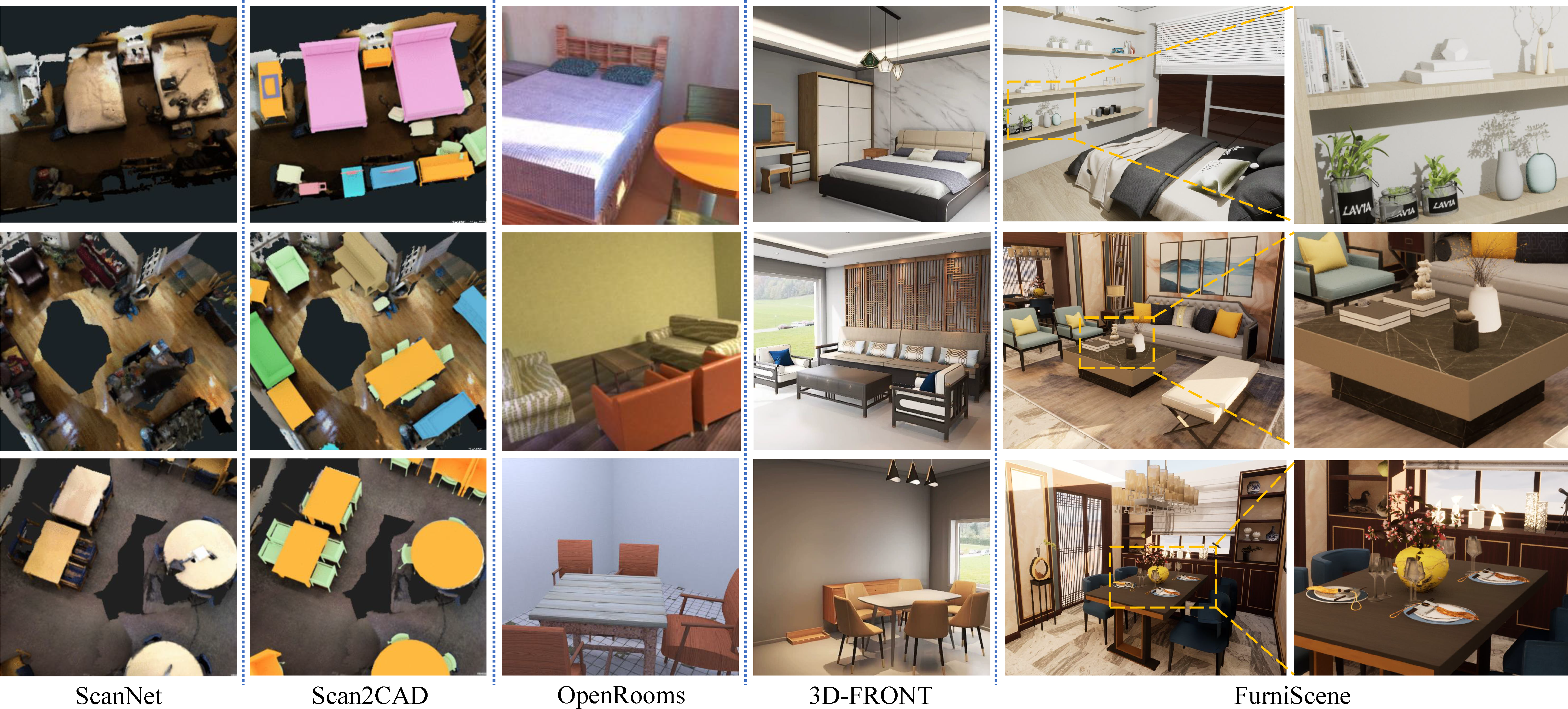}
  \vspace{-0.2cm}
  \caption{Visualization of ScanNet \cite{ScanNet}, Scan2CAD \cite{Scan2CAD}, OpenRooms \cite{OpenRooms}, 3D-FRONT \cite{3D-FRONT}, and FurniScene. ScanNet contains a wealth of ornaments, but it does not separate the mesh. The scene in 3D-FRONT is complete, but there are no small ornaments; FurniScene is diverse, rich in detail, and has a lot of ornaments.}
  \vspace{-0.4cm}
  \label{fig:compare}
\end{figure*}

We parameterize each room $R$ as an unordered sequence of $n$ objects, each object contains four attributes: size, class label, location, and rotation:\vspace{-0.1cm}
\begin{align}
\setlength{\abovedisplayskip}{0pt}
\setlength{\belowdisplayskip}{0pt}
R=\{ o_1,o_2,\dots,o_n\},~~~~o_i=(s_i,c_i,l_i,r_i)
\end{align}
where $s_i\in\mathbb{R}^3$, $c_i\in\mathbb{R}^k$, $l_i\in\mathbb{R}^3$, $r_i\in\mathbb{R}^2$ respectively denote the size, class label, location, and rotation of the object $o_i$. The class label $c_i$ is represented as one-hot vectors of the $k$ classes, $r_i$ is the $[sin, cos]$ of rotation angle around the vertical axis, and the $s_i$, $l_i$ is normalized to be $[-1, 1]$ to have the same range to balance their importance.

\subsection{TSDSM: Two-Step Diffusion Scene Model}

\begin{figure}
  \centering
  \includegraphics[width=0.98\linewidth]{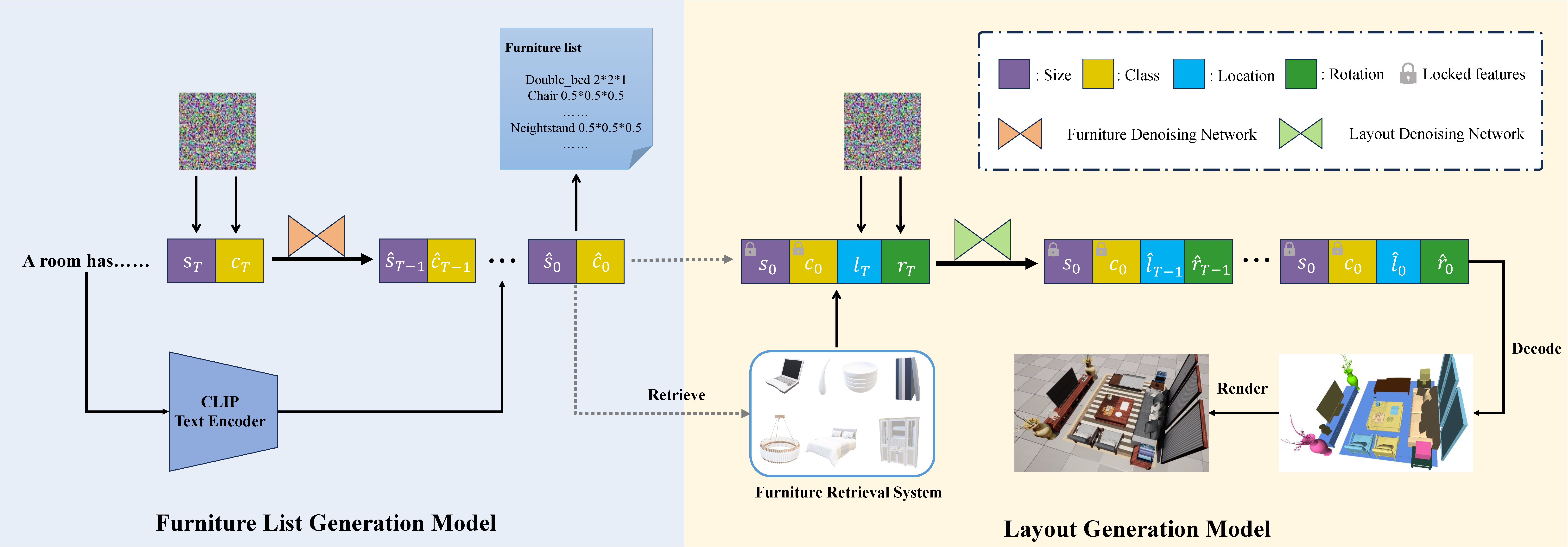}
  \caption{Model architecture. Firstly, FLGM generates a furniture list based on text prompt. Subsequently, FRS utilizes this furniture list to retrieve corresponding furniture models from our furniture database, which are then passed to LGM for the generation of layout information for these models.
  }
  \vspace{-0.4cm}
  \label{fig:method}
\end{figure}

\textbf{Overview}~~As shown in \cref{fig:method}, TSDSM is a two-stage method for indoor scene generation using the diffusion model. In the first stage, the furniture list generation model (FLGM) generates a comprehensive furniture list based on text prompt $text$, specifying the size $\hat{s}$ and category $\hat{c}$ of each furniture item in the room. In the second stage, using the furniture list generated in the first stage, the furniture retrieval system (FRS) can identify the correct furniture items and, based on these items, layout generation model (LGM) generates their corresponding layout information.\vspace{-0.1cm}
\begin{align}
\setlength{\abovedisplayskip}{0pt}
\setlength{\belowdisplayskip}{0pt}
    &\hat{s},\hat{c}=FLGM(text)\\
    &s,c=FRS(\hat{s},\hat{c})\\
    &\hat{l},\hat{r}=LGM(s,c)
\end{align}

\textbf{Furniture List Generation Model}~~To generate the furniture list, we use a diffusion model called FLGM. During the diffusion process, it gradually injects noise to $s_0,c_0$ to obtain noised $s_t,c_t$, where $t\in \{1,\cdots,T\}$ (T=2000 in our work). In the denoising process, we initialize the $s_T, c_T$ using random noise and apply furniture denoising network $\epsilon_{FDN}$ progressively denoise noise into $\hat{s}_0, \hat{c}_0$. The $\epsilon_{FDN}$ takes the time step t, text feature, and $s_t, c_t$ as input and outputs the predicted $\hat{\epsilon}_{s,c}$.\vspace{-0.1cm}
\begin{align}
\setlength{\abovedisplayskip}{0pt}
\setlength{\belowdisplayskip}{0pt}
    \hat{\epsilon}_{s,c}=\epsilon_{FDN}(s_t,c_t,t|text)
\end{align}
The optimization objective of the first stage is the $L_2$ loss for the noise in both size and category $\mathcal{L}_{s,c}$, which can be formally defined as follows:\vspace{-0.1cm}
\begin{align}
\setlength{\abovedisplayskip}{0pt}
\setlength{\belowdisplayskip}{0pt}
    \mathcal{L}_{s,c}=\mathbb{E}_{s_0,c_0,\epsilon_{s,c},t}[||\epsilon_{s,c}-\hat{\epsilon}_{s,c}||^2]
\end{align}

\textbf{Layout Generation Model}~~Once the furniture list is obtained, FRS retrieves the relevant furniture objects from the furniture CAD model database. Then we parameterize the retrieved furniture's size and category information and combine it with random noise to serve as input for the layout denoising network. The diffusion process and denoising process are similar to the furniture list generation model. Importantly, the category and size information remains fixed during this stage and does not participate in the diffusion, denoising process, or loss calculation. The layout denoising network $\epsilon_{LDN}$ takes the time step t, locked $s_0,c_0$, and $l_t, o_t$ as input and outputs the predicted $\hat{\epsilon}_{l,r}$.
\begin{align}
\setlength{\abovedisplayskip}{-5pt}
\setlength{\belowdisplayskip}{-5pt}
    \hat{\epsilon}_{l,r}=\epsilon_{LDN}(lock(s_0,c_0),l_t,r_t,t)
\end{align}
In order to train the LDN, we define two losses, $\mathcal{L}_{l,r}$ to supervise locations and rotation, and $\mathcal{L}_{box}$ to prevent overlapping of objects.\vspace{-0.1cm}
\begin{align}
\setlength{\abovedisplayskip}{-5pt}
\setlength{\belowdisplayskip}{-5pt}
    &\mathcal{L}_{l,r}=\mathbb{E}_{l_0,r_0,\epsilon_{l,r},t}[||\epsilon_{l,r}-\hat{\epsilon}_{l,r}||^2]\\
    &\mathcal{L}_{box}=\sum\limits_{t=1}\limits^{T}w_t*\sum\limits_{o_i,o_j\in\tilde{x}^t_0}IoU(o_i,o_j)
\end{align}
where $o_i,o_j$ represent the rotated bounding boxes of any two objects in clean room $\tilde{x}^t_0$ reconstructed based on $\hat{\epsilon}_{s,c}, \hat{\epsilon}_{l,r}$, and $w_t$ is the weight of the IOU loss at time step $t$.

\begin{figure*}[h]
  \centering
  \includegraphics[width=0.98\linewidth]{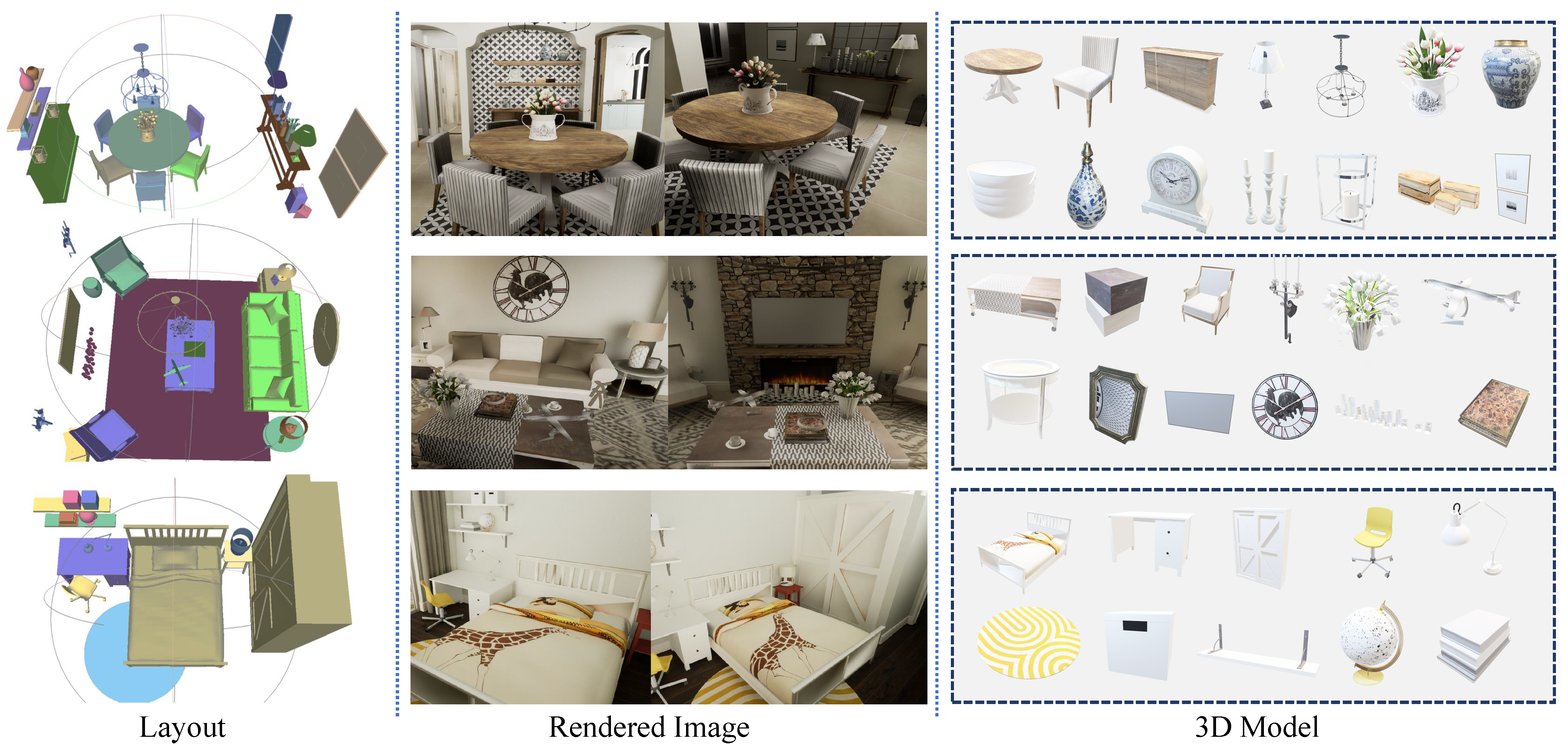}
  \vspace{-0.2cm}
  \caption{Examples of rooms in FurniScene. The left column displays the unrendered complete layout information of the room. The middle column showcases multiple rendered images of the room after UE rendering. The right column displays the CAD models contained in the room, each of which boasts intricate geometric details. It is worth noting that FurniScene includes numerous furnishings such as books, bottles, clocks, vases, \textit{etc.}
 }
 \label{fig:example}
\end{figure*}

\textbf{Denoising Network}~~FDN firstly encodes the features of each piece of furniture using an MLP and treats them as distinct tokens input into a transformer encoder without any positional encoding~\cite{transformer}. Self-attention is applied between different tokens to model the co-occurrence relationships among different pieces of furniture. To inject text features into the feature space, cross-attention is performed between the text feature and the furniture feature. For the LDN, we adopt the UNet-1D structure from DiffuScene~\cite{DiffuScene}.

\section{Experiments}
\label{sec:experiments}
\subsection{Experiment Setup}
\begin{table*}[h]
  \setlength\tabcolsep{5.5pt}
  \centering
  \resizebox{!}{1.6cm}{
  \begin{tabular}{cccccc|cccc|cccc}
  \hline
    \toprule
    \multirow{2}{*}{Method} & & \multicolumn{4}{c}{Bedroom} & \multicolumn{4}{c}{Living Room} & \multicolumn{4}{c}{Dining Room}  \\
     & & FID$\downarrow$ & KID$\downarrow$ & SCA$\%$ &  CKL$\downarrow$ & FID$\downarrow$ & KID$\downarrow$ & SCA$\%$ &  CKL$\downarrow$ & FID$\downarrow$ & KID$\downarrow$ & SCA$\%$ &  CKL$\downarrow$  \\
    \midrule
    SceneFormer~\cite{SceneFormer} & & 85.01 & 23.61 & 95.28 &  289.62 & 91.85 & 36.65 & 98.40 &  547.12 & 120.67 & 30.43 & 93.84 &  301.00\\
    ATISS~\cite{ATISS} & & 82.84 & 41.15 & 91.17 &  \textbf{10.18} & 63.63 & 29.46 & 84.01 &  43.28 & 84.05 & 38.64 & 93.58 &  \textbf{9.01}\\
    DiffuScene~\cite{DiffuScene} & & 110.00 & 73.26 & 93.22 &  82.86 & 74.31 & 46.20 & 85.70 &  72.97 & 115.08 & 66.84 & 93.44 &  87.84\\
    \midrule
    TSDSM (w/o text prompt) & & \textbf{46.37} & 15.34 & 80.82 &  125.42 & 44.45 & 22.65 & \textbf{78.31} &  \textbf{38.30} & \textbf{69.24} & 26.95 & 86.01 &  29.46\\
    TSDSM (w/ text prompt) & & 50.88 & \textbf{14.97} & \textbf{79.95} &  53.44 & \textbf{36.54} & \textbf{19.31} & 85.43 &  49.93 & 75.60 & \textbf{22.69} & \textbf{80.09} &  69.34\\
    \bottomrule
  \end{tabular}
  }
  \vspace{-0.2cm}
  \caption{Indoor scenes generate benchmark performance. For FID, KID, and CKL, lower values indicate more realistic generated scenes. For SCA, values closer to 50 indicate that the generated scenes are more difficult to distinguish from real scenes.}
  \vspace{-0.4cm}
  \label{tab:result}
\end{table*}

\begin{figure*}[h]
  \centering
  \includegraphics[width=0.98\linewidth]{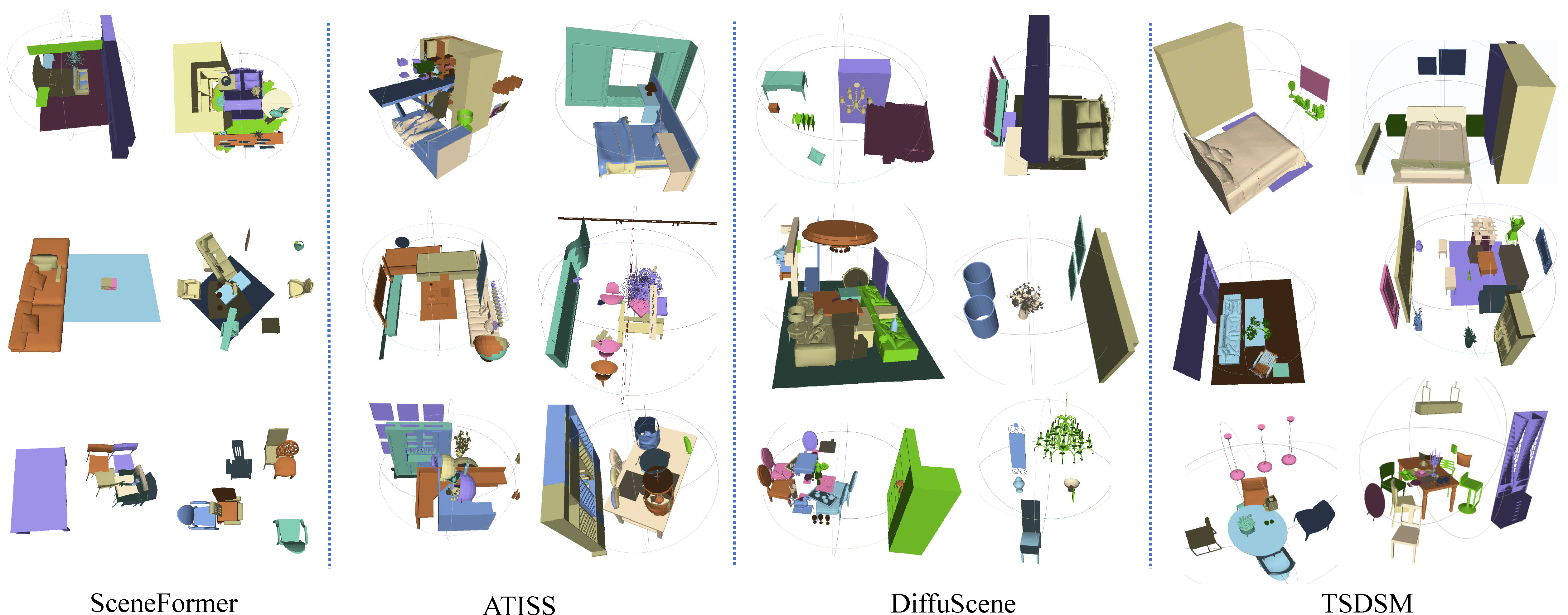}
  \vspace{-0.2cm}
  \caption{Qualitative results of unconditional scene generation. The top row represents the generated results for bedrooms, the middle row corresponds to living rooms, and the bottom row depicts dining rooms.}
  \vspace{-0.4cm}
  \label{fig:prediction}
\end{figure*}

\textbf{Baselines.}~~We use FurniScene to conduct benchmark experiments on the indoor scene generation task based on layout information, testing several methods: 1) SceneFormer~\cite{SceneFormer}, which uses four different transformers to generate the location, size, category, and orientation of the next object to be placed respectively; 2) ATISS~\cite{ATISS}, which uses a transformer encoder and an MLP decoder to autoregressively generate the next furniture property to be placed; 3) DiffusScene~\cite{DiffuScene}, using Diffusion Model to denoise the scene graph from random noise.

\textbf{Evaluation Metrics}~~Following the previous work~\cite{ATISS,DiffuScene}, we employ several established evaluation metrics to measure the authenticity and diversity of 1,000 synthesized scenes, including frechet inception distance (FID), kernel inception distance (KID$\times$0.001), scene classification accuracy (SCA), and category KL Divergence (CKL$\times$0.001).

\textbf{Implementation}~~ We train a separate model for each room type. The training process is conducted on a single NVIDIA A40 GPU with a batch size of 64 and a total of 1000 epochs. The learning rate is initialized at 2e-5 and decreases by 0.5 every 100 epochs. For the diffusion process, we use the original DDPM~\cite{ddpm} settings.

\subsection{Experimental Results}
\textbf{Quantitative Results}~~\cref{tab:result} shows the quantitative comparison on FurniScene. Note that for our method, we have evaluated both the results of text-conditioned generation and unconditional generation. Our method outperforms others, achieving the best metrics on FID, KID, and SCA. These results affirm the high realism exhibited by the scenes generated through our approach. Additionally, our method performs admirably on CKL, indicating that the object categories within the generated scenes closely align with the dataset distribution. In summary, our method achieves the best results in complex scenes with a significant number of furniture items.

\textbf{Qualitative Results}~~\cref{fig:prediction} visually presents the performance of various scene generation methods in bedrooms, living rooms, and dining rooms. It is evident that SceneFormer generates a limited set of categories, with a concentration on objects that frequently appear in specific scenes. Additionally, it exhibits a notable issue of severe object overlap. ATISS generates more plausible category distributions, it frequently produces objects with identical positions, categories, and sizes, leading to significant overlap issues. DiffuScene generates less reasonable category distributions and often results in rooms with either an excessive or insufficient number of objects. Our method, on the other hand, generates furniture categories and quantities with greater plausibility, resulting in more realistic scenes. However, it is unfortunate that object positions and orientations are not optimal.
\vspace{-0.2cm}

\begin{table}[h]
  \centering
  \resizebox{!}{0.9cm}{
  \begin{tabular}{cccccc}
  \hline
    \toprule
    Method & &  FID$\downarrow$ & KID$\downarrow$ & SCA$\%$ &  CKL$\downarrow$\\
    \midrule
    first w/ pos & & 53.53/52.47 & 26.57/21.76 & 85.10/87.75 &  61.44/102.21\\
    second single-head & & 51.54/50.49 & 23.12/21.12 & 82.15/78.28 &  43.91/53.51\\
    second w/o separate & & 100.10/48.82 & 24.62/21.30 & 83.33/80.77 &  42.90/65.22\\
    \midrule
    final & & \textbf{44.45}/\textbf{36.54} & \textbf{22.65}/\textbf{19.31} & \textbf{78.31}/\textbf{85.43} & \textbf{38.30}/\textbf{49.93}\\
    \bottomrule
  \end{tabular}
  }
  \vspace{-0.2cm}
  \caption{Quantitative ablation studies on the task of indoor scene generation on the FurniScene living rooms (w/ text vs. w/o text).}
  \vspace{-0.2cm}
  \label{tab:ablation}
\end{table}

\textbf{Ablation}~~We conduct detailed ablation studies to verify the effectiveness of each design in our TSDSM, with quantitative results provided in \cref{tab:ablation}. We conducted three sets of ablation experiments on indoor scene generation: 1) first w/ pos, which involves adding positional encoding in the first stage; 2) second single-head, which uses single-head transformers in the second stage; 3) second w/o separate, which doesn't apply separate MLPs to different attributes in the second stage. Our findings indicate that adding positional encoding in the first stage is ineffective because our data is unordered. Due to the significant differences in data distributions between different attributes, using single-head transformer or not adding separate MLPs for each attribute in the second stage will result in poorer performances.

\textbf{Analysis}~~SceneFormer's category distribution is poor, but its realism performance is relatively good. This may be attributed to its treatment of indoor scenes as ordered sets, affecting the modeling of furniture long-tail distribution. Compared to SceneFormer and DiffuScene, ATISS demonstrates superior overall performance. This can be attributed to the effective modeling of relationships between different objects by the transformer, as well as the utilization of distinct decoders for each attribute in ATISS. DiffuScene excels in simple scenes but struggles in complex scenes in our FurniScene dataset, possibly due to its single-step attribute generation strategy during network training. Our method outperforms others because the furniture list generated in the first stage provides a strong prior for the layout generation in the second stage. Additionally, as demonstrated in \cref{fig:compare,fig:example}, rooms in FurniScene showcase a greater variety and quantity of objects compared to the previous dataset. This may be the reason for the unsatisfactory performance of existing methods.

\section{Conclusion}
\label{sec:conclusion}

In this paper, we propose FurniScene, a large-scale 3D rooms dataset with intricate furnishing scenes from interior design professionals.  FurniScene surpasses the data scale of all existing datasets, encompassing a large number of rooms with rich details. These meticulously crafted spaces are replete with decorative elements, imbuing them with a high level of authenticity. FurniScene holds the potential to offer enhanced support for subsequent research in indoor scene generation and understanding. Simultaneously, we propose TSDSM, a two-stage diffusion scene model. By dividing the indoor scene generation process into two stages, we address inherent optimization challenges observed in single-stage generation approaches. Experimental results demonstrate that this two-stage model enables the generation of high-quality indoor scenes.
{
    \small
    \bibliographystyle{ieeenat_fullname}
    \bibliography{main}
}



\end{document}